\documentclass[submission,copyright,creativecommons]{include/latex-class-eptcs/eptcs}
\usepackage[utf8]{inputenc}
\usepackage{amsmath}
\usepackage{listings}

\newcommand{\HT}{\ensuremath{\mathrm{HT}}}

\newcommand{\LTL}{\ensuremath{\mathrm{LTL}}}
\newcommand{\LTLf}{\ensuremath{\mathrm{LTL}_{\!f}}}

\newcommand{\THT}{\ensuremath{\mathrm{THT}}}

\newcommand{\TEL}{\ensuremath{\mathrm{TEL}}}
\newcommand{\TELf}{\ensuremath{{\TEL}_{\!f}}}

\newcommand{\LDL}{\ensuremath{\mathrm{LDL}}}
\newcommand{\LDLf}{\ensuremath{\mathrm{LDL}_{\!f}}}

\newcommand{\DHT}{\ensuremath{\mathrm{DHT}}}

\newcommand{\DEL}{\ensuremath{\mathrm{DEL}}}
\newcommand{\DELf}{\ensuremath{{\DEL}_{\!f}}}

\newcommand{\DL}{\ensuremath{\mathrm{DL}}}
 \RequirePackage{bm}
\RequirePackage{textcomp}
\RequirePackage{upgreek}

\IfFileExists{outline.tex}{\input{outline}}{}

\newcommand{\alwaysF}{\ensuremath{\square}}

\newcommand{\eventuallyF}{\ensuremath{\Diamond}}

\IfFileExists{outline.tex}
        {}
        {}
\IfFileExists{outline.tex}
        {}
        {}
\IfFileExists{outline.tex}
        {}
        {}

\IfFileExists{outline.tex}
        {}
        {}

\newcommand{\stp}{\ensuremath{\uptau}}

\newcommand{\dalways}[1]{\ensuremath{[#1]\,}}                        \newcommand{\deventually}[1]{\ensuremath{\langle#1\rangle\,}}

 \newcommand{\sysfont}{\textit}

\newcommand{\asprilo}{\sysfont{asprilo}}

\newcommand{\clingo}{\sysfont{clingo}}

\newcommand{\gringo}{\sysfont{gringo}}

\newcommand{\telingo}{\sysfont{telingo}}

\providecommand{\Underscore}{\textunderscore}

\lstdefinelanguage{clingo}{basicstyle=\ttfamily,keywordstyle=[1]\bfseries,keywordstyle=[2]\bfseries,keywordstyle=[3]\bfseries,showstringspaces=false,literate={_}{\Underscore}1 {\%\%}{}0,escapeinside={\#(}{\#)},alsoletter={\#,\&},keywords=[1]{not,from,import,def,if,else,elif,return,while,break,and,or,for,in,del,and,class,with,as,is,yield,async},keywords=[2]{\#const,\#show,\#minimize,\#base,\#theory,\#count,\#external,\#program,\#script,\#end,\#heuristic,\#edge,\#project,\#show,\#sum},keywords=[3]{&,&dom,&sum,&diff,&show},morecomment=[l]{\#\ },morecomment=[l]{\%\ },morestring=[b]",stringstyle={\itshape},commentstyle={\color{darkgray}}}

\lstdefinelanguage{python}{basicstyle=\ttfamily,keywordstyle=[1]\bfseries,showstringspaces=false,literate={_}{\Underscore}{1},escapeinside={\#(}{\#)},alsoletter={\#,\&},keywords=[1]{not,from,import,def,if,else,elif,return,while,break,and,or,for,in,del,and,class,with,as,is,yield,async},morecomment=[l]{\#\ },morestring=[b]",stringstyle={\itshape},commentstyle={\color{darkgray}}}

\newcommand{\Next}{{\ensuremath{\bigcirc }}}

\newcommand{\eqdef}{\ensuremath{\mathbin{\raisebox{-1pt}[-3pt][0pt]{$\stackrel{\mathit{def}}{=}$}}}}

\newcommand{\s}[1]{\ensuremath{q_{#1}\,}}

\newcommand{\last}{\ensuremath{\mathit{last}}}

\newcommand{\AFW}{\ensuremath{\mathrm{AFW}}}

\newcommand{\handt}{\ensuremath{\langle H,T \rangle}}
\newcommand{\tandt}{\ensuremath{\langle T,T \rangle}}

 \usepackage{tikz} \usepackage{amssymb}
\usepackage{subcaption}
\usetikzlibrary{patterns,shapes.arrows,arrows,automata,positioning,trees,calc,shadows,positioning}
\usetikzlibrary{fit}
\providecommand{\event}{ICLP DC 2021} 

\title{Automata Techniques for Temporal Answer Set Programming}
\author{Susana Hahn
\institute{University of Potsdam, Germany}
\email{hahnmartinlu@uni-potsdam.de}
}   
\def\titlerunning{Automata techniques for temporal ASP}
\def\authorrunning{S. Hahn}
\begin{document}
\definecolor{stylered}{HTML}{F5D5CB}
\definecolor{stylegreen}{HTML}{D7ECD9}
\definecolor{styleyellow}{HTML}{F6F6EB}
\definecolor{stylepurple}{HTML}{D5D6EA}
\maketitle

\begin{abstract}
    Temporal and dynamic extensions of Answer Set Programming (ASP) have played an important role in addressing dynamic problems, as they allow the use of temporal operators to reason with dynamic scenarios in a very effective way. 
    In my Ph.D. research, I intend to exploit the relationship between automata theory and dynamic logic to add automata-based techniques to the ASP solver $\clingo$ helping us to deal with theses type of problems.
\end{abstract} \section{Introduction}\label{sec:introduction}
Representing and reasoning about dynamic domains is a key problem in the field of Knowledge Representation and Reasoning. 
Dynamic and temporal logics are used to describe ordered events, thus they have been adopted as a powerful tool to handle domains where we need to capture actions and change.

While most of the research around these formalisms is grounded on classical logic, there is a growing interest to incorporate such dynamic specifications to reason in a non-monotonic manner. 
One of the main candidates for modeling and solving problems with this type of logic is Answer Set Programming (ASP) \cite{breitr11a}. 
ASP is a well-established approach to declarative problem solving where problems are encoded in the form of logic programs.
In the last years, theoretical work on extending the ASP semantics to handle temporal logics has been perused, helping to enrich the modeling power of ASP. 

A common strategy for addressing well-known problems within these temporal formalisms is an automata-theoretic approach with the underlining idea of constructing an automaton that accepts exactly the models of the formula. 
Nonetheless, there is a lack of research for applying such methods in ASP.
The goal of my Ph.D. project is to employ techniques from Automata Theory in temporal ASP using the ASP system \clingo\ to enhance the language with dynamic constructs that will rely on the computation of finite state machines. 
The hypothesis is that exploiting techniques from such a well-known and highly investigated field will also enrich the performance of temporal systems that use ASP to solve dynamic problems. Even though the primary goal is to explore automata-based practices, the overall aim of the group research project is to extend ASP into a general-purpose technology for dynamic domains. \section{ Background}

A way to formally describe the declarative paradigm of ASP is using Equilibrium Logic (EL) \cite{pearce06a}. 
The \emph{equilibrium models} of this logic are equivalent to the original definition of stable models semantics by  Gelfond and Lifschitz \cite{gellif88b}, but are obtained by enforcing a minimality condition on the logic of here-and-there (HT). 
HT logic, first introduced in \cite{heyting30a}, is a constructive monotonic logic where an interpretation is a pair $\handt$ with $H\subseteq~T$, where atoms in $H$ (here) are \emph{proven} while atoms in $T$ (there) are \emph{assumed} and those not in $T$ are \emph{false}.
Based on this, HT can be used for representing and analyzing logic programs under stable model semantics \cite{pearce96a,ferlif05a} by selecting models that are said to be in equilibrium.
A total model $\tandt$ is said to be in equilibrium iff there is no other $H\subset T$ such that $\handt$ is a also an HT-model.
Under this condition, we obtain a non-monotonic entailment relation where conclusions can be retracted with new information, providing a purely logical characterization for negation as failure.

The quest to extend logic programs with modal temporal operators started in the 1980s under the name of Temporal Logic Programming (TLP) \cite{moszkowski86a,fukotamo86a,gabbay87b,abaman89a,baudinet92a,orgwad92a}, usually based on the classical modal logic Linear-time Temporal logic (LTL). 
LTL is one of the most commonly used temporal logics and it allows to express temporal properties using temporal modal operators such as $\Next$ (next), $\eventuallyF$ (eventually) and $\alwaysF$ (always).  
LTL can also be defined in terms of Dynamic Logic (DL) \cite{hatiko00a}.
DL allows for expressing regular expressions over (infinite) traces, and thus to mix declarative with procedural specifications. 
It also inspired the creation of the action language GOLOG \cite{lereleli97a} based on situation calculus.
Originally, temporal formalisms were investigated for infinite traces, however, in the past decade, the case of finite traces $\cdot_f$ has gained interest, as it can be better fitted to many AI applications and constitutes a computationally more feasible version. 
The introduction of $\LTLf$ and Linear Dynamic Logic over finite traces (\LDLf) \cite{giavar13a} served as a stepping stone to define the syntax and classical semantics under this restriction.

TLP was more recently revised after the appearance of ASP, where the idea was to extend the equilibrium models of $\HT$ to deal with dynamic scenarios. 
The research started with infinite traces giving rise to (Linear) Temporal Here-and-There (\THT) \cite{agcadipevi13a} and (Linear) Dynamic logic of Here-and-There (\DHT) \cite{bocadisc18a} with their non-monotonic counterpart for temporal stable models Temporal Equilibrium Logic (\TEL) \cite{agcadipevi13a} and Dynamic Equilibrium Logic (\DEL) \cite{cadisc19a}. 
These logics build upon the logic of HT together with $\LTL$ \cite{pnueli77a} and $\LDL$\cite{giavar13a}, respectively. 
The idea behind these temporal formalisms is to capture time as sequences of HT-interpretations, called HT-traces, resulting in an expressive non-monotonic modal logic that shares the general syntax of LTL while possessing a computational complexity beyond LTL.
Later on, motivated by how finite traces can better represent the way in which ASP is used to solve problems, the investigation shifted to finite traces with $\TELf$ \cite{cakascsc18a}. 
The temporal operators and semantics of $\TELf$ were introduced into the ASP system \clingo\ enriching its modeling power and yielding the first temporal ASP solver \telingo \cite{cakamosc19a}.
Subsequent work in $\telingo$ incorporated dynamic operators from $\DELf$ \cite{cadisc19a, cadilasc20a} by unfolding their definitions into $\TELf$ with the help of auxiliary predicates.

Several computational approaches for tackling challenges that emerge from temporal formalisms often rely on translations of temporal formulas into an automaton that accepts exactly the models of the formula. 
This relationship has been exploited in different areas of temporal reasoning such as satisfiability checking, model checking and synthesis \cite{vardi95a, vardi97a, giavar15a, zhpuva19a}. 
Furthermore, the field of planning has benefited from temporal reasoning to express goals and preferences using an underlining automaton \cite{bafrbimc08a, giarub18a, baimci06a}.
Given that planning problems are normally encoded in Planning Domain Definition Language (PDDL) \cite{mcdermott98a}, transformations from temporal formalisms to PDDL \cite{crecod04a} as well as from PDDL into ASP \cite{gekaknsc11a} have been developed to bridge this gap.
When constructing an automaton to represent the formula its transitions are subject to the interpretation at the current time point of the trace. 
In the case of infinite traces, automaton on infinite words known as Büchi Automata \cite{buchi90a} have acted as the target structure of the translations.
On the other hand, Non-deterministic finite automata (NFA) \cite{hopull79a} and Deterministic Finite Automata (DFA) work nicely with finite traces with the disadvantage of being of exponential size wrt. the input formula. 
To tackle this issue, M. Vardi and G. De Giacomo proposed a translation \cite{giavar13a} from $\LTLf$ and $\LDLf$ to Alternating Automaton on Finite Words (AFW) \cite{chkost81a,vardi97a,giavar13a} which is an adaptation of Alternating Büchi Automata to finite traces and can be seen as an extension of NFAs by universal transitions.  
This type of automata has also been used as a structure for learning $\LTLf$ formulas \cite{cammci19a}.

On the implementation side, there exists several tools that exploit this mapping. 
For infinite traces, Spot \cite{dulefamirexu16a} is the most widely used system which allows manipulation of $\LTL$ and $\omega$-automata.
Working with finite traces, the system called LTLf2DFA \footnote{\href{https://github.com/whitemech/LTLf2DFA}{https://github.com/whitemech/LTLf2DFA}} translates $\LTLf$ formulas to DFAs using MONA \cite{hejejoklparasa95a}, a tool to translate Monadic Second Order (MSO) formulas into finite-state automata.
Moreover, those systems paved the way for developing of \texttt{abstem} \cite{cabdie14a}, a framework that provides functionalities for temporal theories under $\TEL$ semantics.

Despite the clear link between these areas, there have only been a few attempts to represent temporal and procedural knowledge in ASP using an automata-like definition \cite{sobanamc03a} \cite{ryan14a}.   
Notably, the use of automata encodings in ASP to restrict plans in dynamic domains is an interesting topic to consider.  
This can be studied using the ASP system \clingo\ \cite{gekakasc17a} to extend ASP with dynamic constructs that will rely on the computation of finite state machines. 
With this idea in mind, the intention of the project is to take advantage of the two methods provided by \clingo\ for capturing new functionalities \cite{kascwa17a}: meta programming, which uses the reification feature of \clingo\ allowing us to express the new functionality using ASP; and the sophisticated Python API used to manipulate and customize the internal workflow of the system.
Thanks to such flexibility we will be able to explore multiple ways of representing the different types of automata and their computation as logic programs. 
It will also enable us to alter the propagation inside the solver providing fine-grained handling of the solving process, hence giving us the opportunity to introduce automata-based techniques like the ones previously mentioned.
 \section{Status}

In the current status of the project, we have not yet explored the non-monotonic side of temporal reasoning. 
Even though the semantics we have used so far for the temporal formalisms have been monotonic, we were able to incorporate them in ASP by restricting the dynamic formulas to integrity constraints where their behavior is classical.
With this in mind, currently, we can not conclude a dynamic behavior, however, we can use it to filter plans and perform other tasks as we will mention below.
These initial restrictions allowed us to start our exploration with the translation from $\LDLf$ to $\AFW$ \cite{giavar13a} and to explore other existing tools for obtaining an automata-representation of the dynamic formula.
Our research has also been inclined towards dynamic over temporal formulas since we can represent the formulas in $\LTLf$ using operators from $\LDLf$.

\begin{figure}[h]
    \centering
    \begin{tikzpicture}
\tikzstyle{grid}=[gray,very thin,opacity=0.2]
    \tikzstyle{arrowtext}=[midway,above, fill=white, inner sep=0pt, outer sep=5pt]
    \tikzstyle{approach}=[fill=gray!10,  inner sep=5pt, align=center]
    \tikzstyle{traces}=[font=\footnotesize, fill=stylepurple,  inner sep=5pt, align=center]

    \tikzstyle{element}=[font=\scriptsize, fill=styleyellow, draw=black, inner sep=5pt, align=center]
    \tikzstyle{encoding}=[element, fill=stylered]
    \tikzstyle{input}=[element, fill=styleyellow]
    \tikzstyle{output}=[element, fill=stylegreen]
    \tikzstyle{multioutput}=[output, double copy shadow={shadow xshift=0.5ex,shadow
    yshift=-0.75ex,draw=black!30}]

    \tikzstyle{arrw}=[black!50]
    \tikzstyle{trans}=[font=\tiny]

    \tikzstyle{system}=[double, rounded corners, font=\scriptsize, fill=gray!20, draw=black, inner sep=5pt, align=center]

\def\yresultsize{1}
    \def\ylastcallsize{3}
    \def\yapproachsize{5}
    \def\yinputsize{2}

    \def\yapproachmin{\yresultsize + \ylastcallsize}
    \def\yapproachmax{\yapproachmin + \yapproachsize}

    \def\ytop{\yapproachmax+\yinputsize}
\def\xafwsize{5}
    \def\xmonasize{5}

    \def\xspace{0.8}
    \def\xafwmin{\xspace}
    \def\xafwmax{\xafwmin+\xafwsize}
    \def\xmonamin{\xafwmax+\xspace}
    \def\xmonamax{\xmonamin+\xmonasize}

\node[approach, fit={(\xafwmin,\yapproachmin) (\xafwmax,\yapproachmax)}, label={\footnotesize{AFW using ASP}}] (afw) {};
    \node[approach, fit={(\xmonamin,\yapproachmin) (\xmonamax,\yapproachmax)}, label={\footnotesize{DFA using MONA}}] (mona) {};

\node[approach, fit={(\xmonamin,\yapproachmax+1.4) (\xmonamax,\ytop+.4)}, fill=gray!10] (domain) {};

    \node[encoding] (grammar) at (1.75,\ytop) {\texttt{grammar.lp}};

    \node[input] (constraint) at (4.75,\ytop) {\texttt{<dyncon>.lp}};

    \node[input] (instance) at (8,\ytop) {\texttt{<ins>.lp}};
    \node[input] (encoding) at (10.5,\ytop) {\texttt{<enc>.lp}};

    \node[system] (afwgringo) at (\xafwmin + \xafwsize/2,\yapproachmax-.5) {\gringo};
    \node[output] (reified) at (\xafwmin + 1,\yapproachmax-2) {\texttt{reified.lp}};
    \node[encoding] (ldlf2afw) at (\xafwmin + 4,\yapproachmax-2) {\texttt{ldlf2afw.lp}};
    \node[system] (afwclingo) at (\xafwmin + \xafwsize/2,\yapproachmax-3.5) {\clingo};

    \node[output] (afwautomtaton) at (\xafwmin + \xafwsize/2,\yapproachmin-1) {\texttt{afw.lp}};
    \node[system] (afwfinal) at (\xafwmin + \xafwsize/2 +1 ,\yapproachmin-2.5) {\clingo};

    \node[system] (monaclingo) at (\xmonamin + \xmonasize/2,\yapproachmax-.5) {\clingo API};
    \node[encoding] (monaldlf) at (\xmonamin + \xmonasize/2 ,\yapproachmax-1.5) {\texttt{ldlf2mso.py}};
    \node[multioutput] (monamso) at (\xmonamin + 1,\yapproachmax-2.5) {\texttt{mso.mona}};
    \node[multioutput] (monastm) at (\xmonamin + 4,\yapproachmax-2.5) {\texttt{mso.mona}};
    \node[system] (monamona) at (\xmonamin + \xmonasize/2,\yapproachmax-3.5) {MONA};
    \node[multioutput] (monadot) at (\xmonamin + \xmonasize/2 ,\yapproachmax-4.5) {\texttt{dfa.dot}};

    \node[output] (monaautomtaton) at (\xmonamin + \xmonasize/2,\yapproachmin-1) {\texttt{dfa.lp}};
    \node[system] (monafinal) at (\xmonamin + \xmonasize/2 -1 ,\yapproachmin-2.5) {\clingo};

    \node[encoding] (run) at (\xafwmax + \xspace/2,\yapproachmin-1) {\texttt{run.lp}};

    \node[text width=10cm, traces, fit={(0,0) (\xmonamax+\xspace,0.5)}] (traces) {\scriptsize{Traces for the dynamic problem (\texttt{<ins>.lp} and \texttt{<enc>.lp}) satisfying the dynamic constraint \texttt{<dycon>.lp}}};

\draw[arrw, -latex] (constraint) -- ++(down:0.5) -- ++(left:1.5) -- ++(down:0.5) -- ++(left:1.5) -- ++(down:1.5)  -- (afwgringo.west);
    \draw[arrw, -latex] (grammar) -- ++(down:0.5) -- ++(right:1.5) -- ++(down:0.5) -- ++(right:3.75) -- ++(down:1.5) -> (monaclingo.west);

    \draw[arrw] (instance.south) -- ++(left:0.5) -- ++(down:0.215) -- ++(left:4.25);

    \draw[arrw, -latex] (afwgringo.south) -- ++(down:0.5) -- ++(left:1.5) -> (reified.north);
    \draw[arrw, -latex] (reified) -- ++(down:1.5) -- (afwclingo.west);
    \draw[arrw, -latex] (ldlf2afw) -- ++(down:1.5) -- (afwclingo.east);
    \draw[arrw, -latex] (afwclingo) -- (afwautomtaton.north);
    \draw[arrw, -latex] (afwautomtaton) -> ++(down:1.5) -> (afwfinal.west);
    \draw[arrw, -latex] (run) -> ++(down:1.5) -> (afwfinal.east);
    \draw[arrw, -latex] (afwfinal) -> ++(down:0.8);

    \draw[arrw, -latex] (monaclingo.south) -> (monaldlf.north);
    \draw[arrw, -latex] (monaldlf) -- ++(left:1.5) node[trans,above] {mso($\displaystyle 0,\varphi $)} -> (monamso.north);
    \draw[arrw, -latex] (monaldlf) -- ++(right:1.5) node[trans,above] {STm($\displaystyle 0,\varphi $)} -> (monastm.north);
    \draw[arrw, -latex] (monastm) -- ++(down:1) -> (monamona.east);
    \draw[arrw, -latex] (monamso) -- ++(down:1) -> (monamona.west);
    \draw[arrw, -latex] (monamona) -> (monadot.north);
    \draw[arrw, -latex] (monadot) -> (monaautomtaton.north);
    \draw[arrw, -latex] (monaautomtaton) -> ++(down:1.5) -> (monafinal.east);
    \draw[arrw, -latex] (run) -> ++(down:1.5) -> (monafinal.west);
    \draw[arrw, -latex] (monafinal) -> ++(down:0.8);

    \draw[arrw] (instance) -- ++(down:0.5) -- ++(right:2.5);
    \draw[arrw] (encoding) -- ++(down:0.5) -- ++(left:1.25) -- ++(down:0.5) -- ++(right:2.75) -- ++(down:7.75) -- ++(left:7.72);
\draw[arrw, latex-] (afwfinal) -> ++(up:.75);
    \draw[arrw, latex-] (monafinal) -> ++(up:.75);

\end{tikzpicture}

     \caption{Workflows of our framework. Elements in yellow correspond to user input regarding the dynamic problem, green ones are automatically generated, and red ones are provided to solve the problem.}
    \label{fig:workflow}
\end{figure}
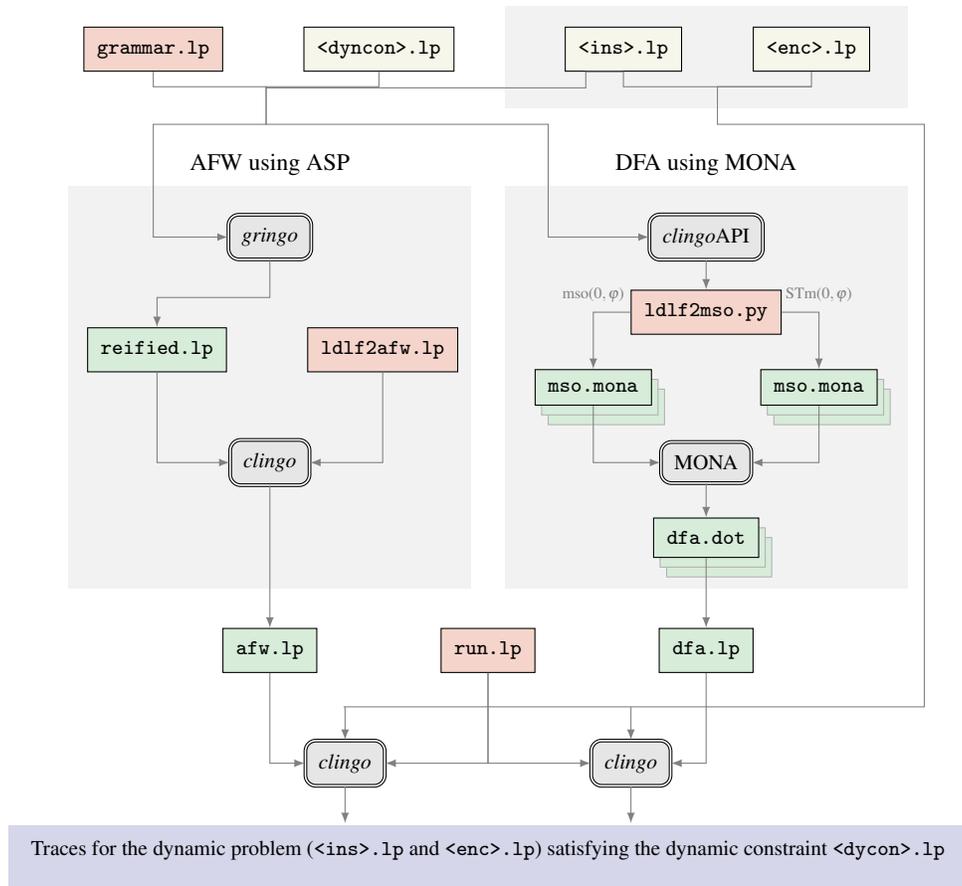

An overview of the current status of the framework \footnote{Git repository with the code of the framework \href{https://github.com/susuhahnml/atlingo}{https://github.com/susuhahnml/atlingo}} can be found in Figure~\ref{fig:workflow}. 
The idea is to compute all fixed-length traces, or plans, of a dynamic problem expressed in
plain ASP (in files \texttt{<ins>.lp} and \texttt{<enc>.lp}) that satisfy the
constraints expressed as dynamic formulas in \texttt{<dyncon>.lp}.
To illustrate its components we will use the following dynamic formula $\varphi$ as an example, showing also its corresponding $\LTLf$ representation \footnote{See \cite{cadilasc20a} as reference for the syntax.}:
\begin{align}\label{eq:main-example}
  \varphi = \deventually{(\dalways{\stp^*} b)? \;}\deventually{ \stp} a \eqdef \alwaysF b \wedge \Next a,
\end{align}
The dynamic version of $\varphi$ can be read as ``there is a path in which, after some number of steps, $b$ holds and after a single step $a$ most hold", while its temporal equivalent is read as ``$b$ always holds and $a$ is true at the next step".
To include these types of formulas in the language, first, we extended $\clingo$'s syntax with a customized theory-specific grammar. 
This newly introduced syntax, which is the same as the one used in the dynamic extension of \telingo \cite{cadilasc20a}, allows us to use the dynamic operators from $\LDLf$ as shown in the dynamic constraint of listing~\ref{lst:theory:del-constraint}.

\begin{lstlisting}[float,label={lst:theory:del-constraint},caption={Integrity constraint for running example (\ref{eq:main-example})},captionpos=b,basicstyle=\ttfamily\footnotesize]
:- not &del{  ? (* &t .>* b) ;; &t .>? a }.

\end{lstlisting}

Providing the grammar, we can use the grounder \gringo\ to output a linearized representation of the formula as facts, such a process is called reification. 
An alternative option is to parse the formula using $\clingo$'s python API and manipulate it.
We will start by talking about the first option (left side of the workflow diagram) which allows us to use solely ASP throughout the implementation.
To do so, once we have the reified output we can use an ASP encoding to process the facts and, in our case, to translate the formula into a declarative representation of the alternating automata, shown in listing~\ref{lst:declarativerep}.
The automaton is represented by predicates: \texttt{prop/2}, providing a symbol table mapping integer identifiers to atoms; \texttt{state/2}, providing states along with their associated dynamic formula; \texttt{initial\_state/1} distinguishing the initial state; \texttt{delta/2}, identifying a possible transition from a state; \texttt{delta/4}, providing conditions on the propositions for the given transition; and \texttt{delta/4}, defining its successor states.

For the second option, which uses $\clingo$'s API (right side of the workflow diagram), we wanted to employ the state-of-the-art tool MONA to obtain a DFA equivalent to the dynamic formula. 
We came up with two different translations from $\LDLf$ to $MSO$.
One of them adapted from \cite{ARHandbook} which translates $\DL$ to First-order infinitary logic, and the other inspired on the definition of \cite{zhpuva19a} for $\LTLf$.
For the implementation, we parsed the formula with $\clingo$'s API and called MONA to obtain the corresponding DFA.
MONA's output was then transformed into facts to use the same declarative structure from Listing~\ref{lst:declarativerep}. 

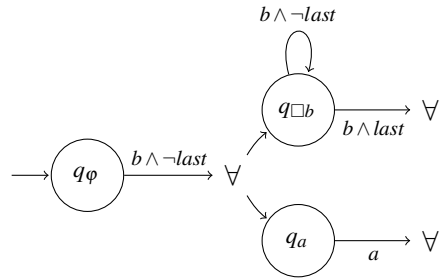
\begin{figure}[h]
    \centering
    \begin{tikzpicture}[shorten >=.8pt,node distance=30pt,on grid,auto,initial text=]
    \tikzstyle{every state}=[font=\footnotesize]
    \tikzstyle{interp}=[font=\scriptsize]
    \tikzstyle{and}=[circle,fill,scale=0.3]

    \node[state,initial]  (s_0) {$\s{\varphi}$};
\node[]            (and)    [right = 55pt of s_0]  {$\forall$};
    \node[state]          (s_a)    [below right = 35pt of and]  {$\s{a}$};
    \node[]           (true_a) [right = 50pt of s_a] {$\forall$};
    \node[state]          (s_b)    [above right = 35pt of and]    {$\s{\alwaysF b}$};
    \node[]          (true_1) [right = 50pt of s_b] {$\forall$};
    
    \path[->]
    (s_0) edge   node[interp] {$b \wedge \neg \last$} (and)
    (and) edge[bend left=10] (s_b)
          edge[bend right=10] (s_a)
    (s_b) edge[below] node[interp] {$b \wedge \last$} (true_1)
          edge[loop above] node[interp] {$b \wedge \neg last$} (s_b)
    (s_a) edge[below] node[interp] {$a$} (true_a);
\end{tikzpicture}     \caption{\AFW\ corresponding to the translation of formula (\ref{eq:main-example}). $\forall$ transitions are universal, meaning that there must be an accepted path form all the outgoing states. The proposition $\mathit{last}$ only holds in the last time point of the finite trace. For more details refer to \cite{giavar13a}. }
    \label{fig:ldlafw}
\end{figure}

\begin{lstlisting}[float,label={lst:declarativerep},caption={Declarative representation of the \AFW\ in Figure~\ref{fig:ldlafw} obtained by translating  formula (\ref{eq:main-example}).},captionpos=b,basicstyle=\ttfamily\footnotesize]
prop(10,"b"). prop(14,"a"). prop(16,"last").
state(0,diamond(sequence(test(box(star(stp),prop(10))),stp),prop(14))).
state(1,prop(14)).
state(2,box(star(stp),prop(10))).
initial_state(0).
delta(0,0).
delta(0,0,out,16). delta(0,0,in,10).
delta(0,0,1). delta(0,0,2).
delta(1,0).
delta(1,0,in,14).
delta(2,0).
delta(2,0,out,16). delta(2,0,in,10).
delta(2,0,2).
delta(2,1).
delta(2,1,in,16). delta(2,1,in,10).
\end{lstlisting}

By having a unified representation to capture the different automata ($\AFW$ and DFA) we were able to craft a single encoding (Listing~\ref{lst:runs}) for checking the runs of any of the mentioned automata. 
We then called \clingo\ with this encoding and the domain-specific knowledge for the dynamic problem such that the output of the call corresponds to the stable models that satisfy the dynamic constraint.

\begin{lstlisting}[float=ht, label={lst:runs}, language=clingo, caption={Encoding defining the accepted runs of an automaton \texttt{run.lp}.},captionpos=b,basicstyle=\ttfamily\footnotesize]
node(Q,0) :- initial_state(Q).%%#(\label{lst:runs:one}#)

1{select(C,Q,T):delta(Q,C)}1 :- node(Q,T), T<=horizon.%%#(\label{lst:runs:two}#)

node(Q',T+1) :- select(C,Q,T), delta(Q,C,Q').%%#(\label{lst:runs:tri}#)

:- select(C,Q,T), delta(Q,C,in,A), not trace(A,T).%%#(\label{lst:runs:for}#)
:- select(C,Q,T), delta(Q,C,out,A), trace(A,T).%%#(\label{lst:runs:fiv}#)

:- node(ID,horizon+1), not final_state(ID).
\end{lstlisting}

 \section{Results}

Considering our experimental studies so far, we explored different dynamic problems such as an elevator moving up and down \cite{cadilasc20a}, the towers of Hanoi, and the blocks world problem.
However, we have mainly focused on the $\asprilo$ \cite{geobotscsangso18a} framework consisting of a versatile benchmark generator, solution checker, and visualizer.
This framework operates in the domain of robotic intra-logistics, where the common setting involves multiple robots, shelves, and stations, placed in a warehouse environment, along with a set of orders.
With this setup, we ran several benchmarks for various instances and constraints to compare the different methods of handling a dynamic constraint.

\begin{figure}
    \centering
\begin{subfigure}{.45\textwidth}
    \centering
    \includegraphics[scale=0.18]{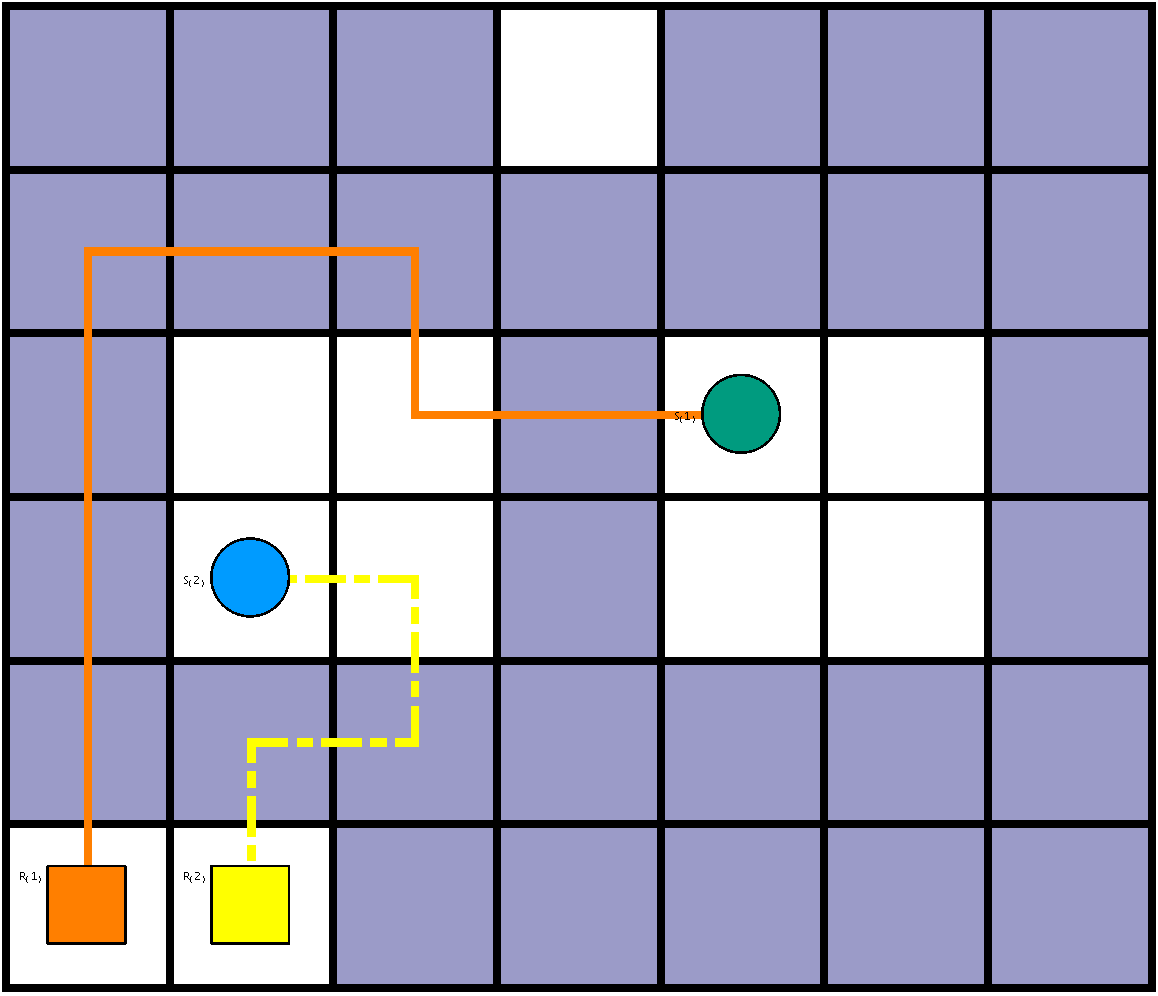}
    \caption{Movement of robots without constraint.}
\end{subfigure}\hspace{10px}
\begin{subfigure}{.45\textwidth}
    \centering
    \includegraphics[scale=0.18]{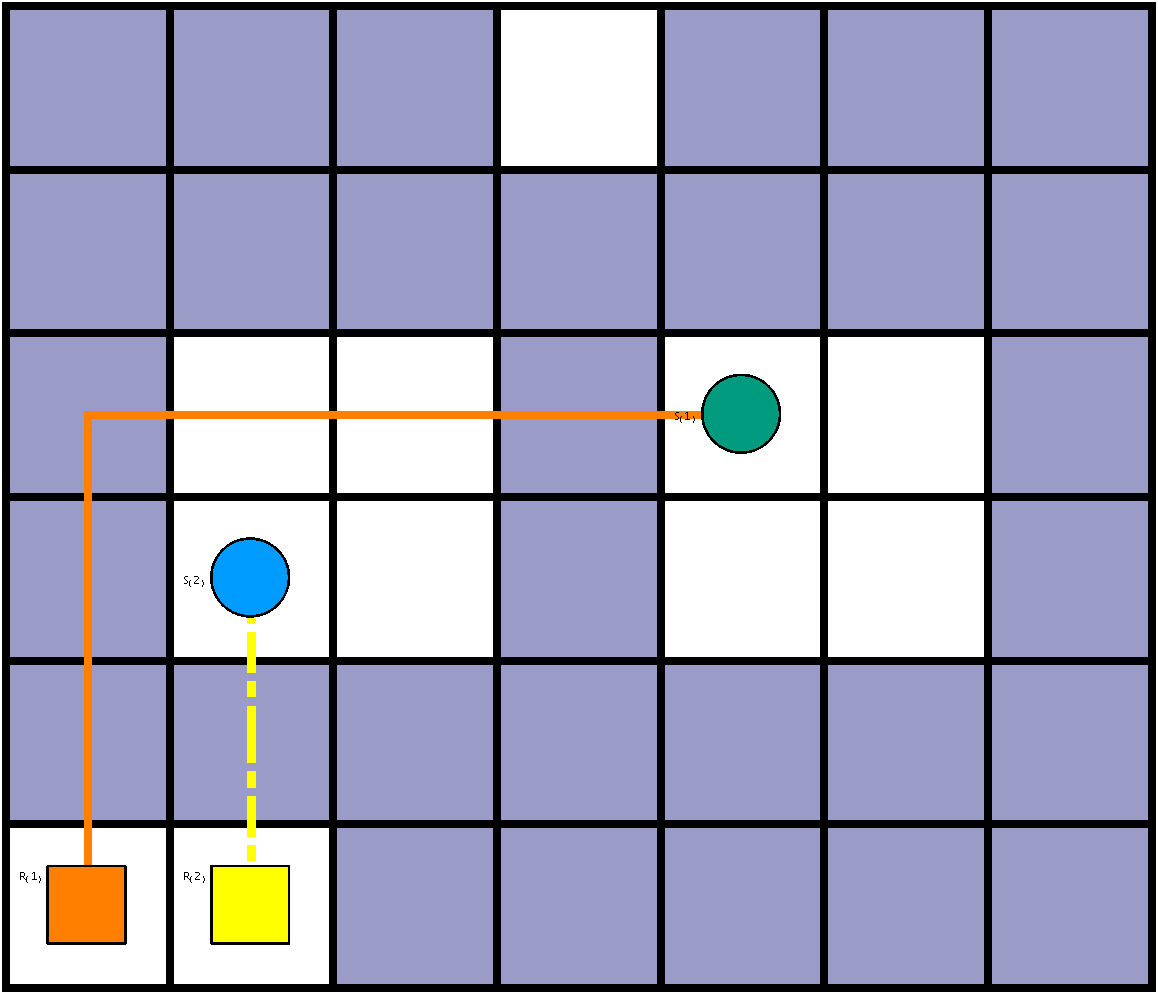}
    \caption{Asprilo visualization for plan following constraint of Listing \ref{lst:asprilo-constraint}.}
\end{subfigure}
\caption{Images from $\asprilo$'s visualizer corresponding to presented plans. Robots are represented by squares and destinations by circles.}
\label{fig:asprilo}
\end{figure}

We illustrate the usage of the framework in \asprilo\ with the constraint of listing~\ref{lst:asprilo-constraint}.
This condition forces all robots to move up many times followed by moves right and then waiting at the destination until the end of the trace.
In figure~\ref{fig:asprilo} we can notice that enforcing this constraint filters out unwanted plans for our simple example.

\begin{lstlisting}[float=ht, label={lst:asprilo-constraint}, language=clingo, caption={Example of dynamic constraint for $\asprilo$.},captionpos=b,basicstyle=\ttfamily\footnotesize]
:- not &del{ *( &t ;; ?move_up(robot(R))) ;;
             *( &t ;; ?move_right(robot(R))) ;; 
             *( &t ;; ?wait(robot(R))) .>? (&t.>* &false) },
        robot(R).
\end{lstlisting}

We have performed experiments using over 8 different dynamic constraints with increasing complexity. 
Our last results showed that including the dynamic constraint reduced the number of decisions that were taken along the search.
We also noticed how, as expected, the size of the automata was reduced by using \AFW\ over DFA. 
However, the performance when the automata were introduced in the solving had no clear winner.
Regarding the translation and preprocessing, we saw that when using more complex dynamic formulas the calls to MONA did not scale, reaching the maximum amount of nodes permitted.
Additionally, we compared these approaches to the dynamic extension of \telingo. 
Overall $\telingo$ showed better results, nonetheless, it must perform the preprocessing for every horizon, whereas the automata representations are independent of the horizon thus they only need to be constructed once.

 \section{Future research}

There remain several paths to explore for this project, as our next steps we have:
\begin{itemize}
    \item The introduction of non-monotonicity in the automata. 
    \item Using an automaton as a Propagator to modify $\clingo$'s solving process.
    \item Use our current setup to decide if there exists a plan where the constraint is satisfied regardless of the horizon by checking if the automaton is empty. For this feature to show interesting results we need to represent the complete dynamic problem as part of the dynamic constraint and tackle complexity issues.
    \item Finally, we intend to investigate other types of automata that can allow more complex behavior such as Timed automata, utilized as modeling formalism for Metric Interval Temporal Logic \cite{menipu06a}.
\end{itemize} 
\bibliographystyle{include/latex-class-eptcs/eptcs}

\end{document}